\documentclass{article}

\usepackage[preprint]{neurips_2026}

\usepackage[utf8]{inputenc} 
\usepackage[T1]{fontenc}    
\usepackage[breaklinks=true]{hyperref} 
\usepackage{url}            
\usepackage{booktabs}       
\usepackage{amsmath}        
\usepackage{amsfonts}       
\usepackage{nicefrac}       
\usepackage{microtype}      
\usepackage{xcolor}         
\usepackage[pdftex]{graphicx}
\usepackage{subcaption}
\usepackage{array}
\usepackage{amsmath}
\usepackage{enumitem}
\usepackage[toc,page]{appendix}
\usepackage{wrapfig}

\usepackage{tikzducks}

\title{The Importance of Encoder Choice: \\A Tabular-Image Study}

\author{%
  Ilia Koloiarov\\
  ISMLL \& VWFS DARC\\
  University of Hildesheim\\
  Hildesheim, Germany \\
  \texttt{koloiarov@ismll.de}
  \And
  Diego Coello de Portugal Mecke\\
  ISMLL \& VWFS DARC\\
  University of Hildesheim\\
  Hildesheim, Germany \\
  \texttt{coello@ismll.de}
  \AND
  Vijaya Krishna Yalavarthi\\
  ISMLL \& VWFS DARC\\
  University of Hildesheim\\
  Hildesheim, Germany \\
  \texttt{yalavarthi@ismll.de}
  \And
  Tom Hanika\\
  ISMLL \\
  University of Hildesheim\\
  Hildesheim, Germany \\
  \texttt{hanika@ismll.de}
  \And
  Lars Schmidt-Thieme\\
  ISMLL\\
  University of Hildesheim\\
  Hildesheim, Germany \\
  \texttt{schmidt-thieme@ismll.de} \\
}

\begin{document}

\maketitle

\begin{abstract}
    Multimodal learning usually requires a dedicated encoder per modality. When a tabular modality is involved, prior work has been mostly using a \emph{plain MLP} as the encoder. Yet if it were a strong encoder, the tabular domain would not be ``the last unconquered castle for deep learning''. This study evaluates state-of-the-art tabular models as encoders in the image-tabular setting for the first time. An obstacle stands out. In-Context Learning models, among the best performing methods in the tabular domain, require labels to process instances, making it non-trivial to embed training and test instances the same way. We addressed this problem across multiple models of this family. With this study, we would like to highlight the importance of encoder factor in the multimodal learning.\footnote{This paper contains color figures. We recommend reading it digitally for the best experience.}
\end{abstract}


\section{Introduction}

The tabular modality is a common component of real-world multimodal datasets:  medical cases link lab values with imaging or timeseries~\citep{Tschandl2018,Dong2020,Johnson2016.MIMICIII}, question-answering systems couple natural language queries with structured records~\citep{Jin2022}, car listings combine metadata with photographs~\citep{Huang2022a}, and art auctions pair provenance data with the painting itself~\citep{WikiArtorgVisualArtWikiArtorgVisualArtEncyclopedia}.

We focus this study on the tabular-image setting. The choice is deliberate. Unlike its text-tabular sibling, where large language models can ingest serialized table representations directly~\citep{Fang2024,Li2025a},  no analogous simple shortcut exists for image-tabular domain. Therefore, a dedicated tabular encoder is required to combine embeddings with image representations.

Yet the literature has almost always used a \emph{plain shallow} MLP~\citep{Wolf2022.DAFT,Hager2023.Best,Taleb2022,Duenias2025} with rare exceptions~\citep{Du2023,Huang2023}. An MLP can perform competitively on tabular data, but only when carefully adapted to the tabular domain~\citep{Kadra2021,Gorishniy2021,Gorishniy2024.TabM,Holzmuller2025.RealMLP}. We assume that a stronger tabular predictor also yields better representations of an isolated encoder for multimodal fusion.

\begin{wrapfigure}{r}{0.35\linewidth}
    \vspace{-1em}
    \center
    \includegraphics[width=\linewidth]{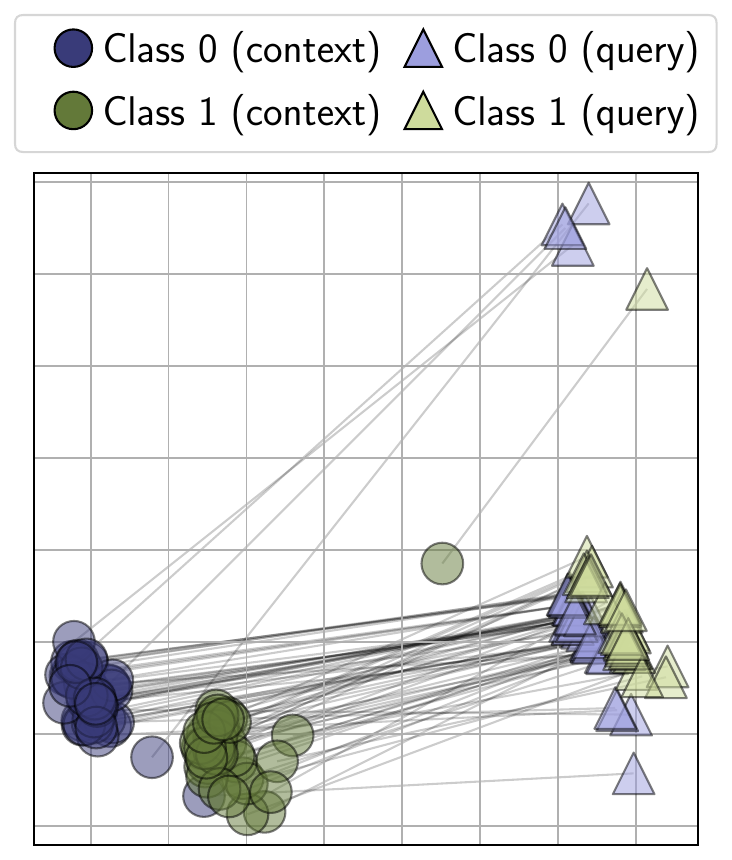}
    \vspace{-1.5em}
  \caption{\small{Context (dark circles) and query (bright triangles) encodings of the same instance (connected by gray lines) do \emph{not coincide} in latent space. 2D PCA of TabPFNv2 on DVM.}}
  \label{fig:train_connected}
  \vspace{-2em}
\end{wrapfigure}

The neglect of the tabular encoder has a historical explanation. For years, gradient-boosted decision trees (GBDTs) dominated tabular benchmarks~\citep{Grinsztajn2022.Why,Borisov2022.Deep,Shwartz-Ziv2022.Tabular}, leaving no compelling Deep Learning candidate for the tabular side of a multimodal pipeline. GBDTs, while strong predictors, do not produce continuous representations compatible with a multimodal pipeline. With no strong tabular encoder available, experimenting with different representations was not a meaningful question.

The recent advances in Deep Learning finally provide strong tabular encoder candidates~\citep{Erickson2025.TabArena,Liu2025.Talent}. Among those, one family has been particularly consequential in shifting the status quo: In-Context Learning Tabular Foundation Models (ICL TFMs)~\citep{Hollmann2023.TabPFN, Hollmann2025.Accurate,Ma2025.TabDPT,Qu2025.TabICL}. They achieve state-of-the-art performance without task-specific retraining, making them natural plug-in components for a multimodal pipeline. 

Using ICL TFMs as tabular encoders, however, is not straightforward. The natural way to embed data with an ICL TFM is to pass training instances as \emph{context} (labels required) and test instances as \emph{queries} (labels unavailable).~\citet{Ye2025.Closer} observed that, in TabPFNv2, this approach causes context and query embeddings to occupy different parts of the latent space. Figure~\ref{fig:train_connected} sharpens the finding: the representation difference persists when the \emph{same instances} appear in both roles simultaneously, ruling out data distributional shift as the cause.  as we show in the paper, this is detrimental for any downstream model. One trained on representations from the context role encounter those from the query role at inference. Therefore, their representations violate the i.i.d.\ assumption. Before our work, \emph{it was unknown} whether this representation shift extends to other ICL TFMs, for instance, TabDPT~\citep{Ma2025.TabDPT} and TabICLv2~\citep{Qu2025.TabICL}.

Taken together, these observations motivate the first systematic sweep over tabular encoders in the image-tabular setting, addressing the ICL TFM representation shift in the process. This yields four findings:

\begin{enumerate}[itemsep=1pt]
  \item Multimodal rankings are not stable across tabular encoders, and conclusions drawn from a single encoder do not generalize. A carefully engineered multimodal method can outperform a simple baseline with a weak encoder, and be matched by the same baseline with a strong one.
  \item Multimodal methods do not consistently outperform unimodal baselines. In datasets where one modality carries little predictive signal, combining modalities can degrade performance below the unimodal baseline, an effect observed in multiple models.
  \item In \emph{likely} multimodal datasets, the observed fusion advantage may be inflated by encoder insufficiency, suggesting that a stronger unimodal encoder yields a more reliable estimate of the true benefit of combining modalities.
  \item On \emph{likely} multimodal datasets, a simple bilinear fusion baseline paired with a strong encoder performs on par with a carefully engineered multimodal method that uses on average 13.2 times more parameters and includes a dedicated pretraining stage.
  \item The context-query representation shift is not specific to TabPFNv2. TabDPT and TabICL exhibit the same behavior, though to a smaller extent. Vanilla($\star$) extraction consistently degrades performance across all three ICL TFMs and should be avoided. NP($\wr$) is the recommended feature extraction for most of the cases.
\end{enumerate}

Each finding is established in the correspondingly numbered subsection of Section~\ref{sec:results}. Code\footnote{\url{https://anonymous.4open.science/r/nips-7aca8573-3137}} and per-trial results are provided as supplementary material. An interactive MLFlow log covering every trial will be released once anonymity constraints are lifted. A preview is already available in the code repository.

\section{Related work}

In this section, we cover the related work most directly relevant to our contributions: tabular in-context learning foundation models, the context-query representation shift, and ICL feature extraction schemes.

\subsection{In-context learning tabular foundation models}

An ICL TFM takes a labeled context set $\mathcal{C} = \{(\mathbf{x}_i, y_i)\}_{i=1}^{n}$ and a query instance $\mathbf{x}$, and produces a prediction $\hat{y}_q = f(\mathbf{x};\, \mathcal{C})$ without weight updates. This paper studies three such models. TabPFNv2~\cite{Hollmann2025.Accurate} tokenizes feature columns and is pretrained on synthetic datasets. TabDPT~\cite{Ma2025.TabDPT} tokenizes rows, is pretrained with self-supervised objectives on real data, and has a context retrieval mechanism. TabICLv2~\cite{Qu2026.TabICLv2} is pretrained on synthetic tabular data with a Set Transformer and an attention temperature scaling method.

\subsection{Context-query representation shift}
\label{sec:icl_shift}

\citet{Ye2025.Closer} first observed that in TabPFNv2, embeddings of context and query instances form separate clusters in the latent space even when there is no intrinsic data distributional shift. Their argument, however, hinges on visual inspection of 2D Principal Component Analysis (PCA) projections~\citep{Pearson1901.Lines} and performance degradation, neither of which is conclusive: the visual separation may be a low-dimensional projection artifact, and performance degradation may stem from the training procedure. We address both limitations in our experiments.

Two structural reasons might explain this: (i) context instances are processed with their labels visible, while queries are not, so the same weights see different inputs depending on the role; (ii) the loss is computed only on query predictions, and nothing in it penalizes the resulting divergence: there is no alignment objective forcing context and query embeddings into a shared space. As the result, the instance representation is determined not only by its features but by the role it was assigned. As a structural property of the encoder, the shift affects any downstream model built on ICL TFM representations. The rest of this paper asks how far it reaches across ICL TFMs, whether it can be mitigated, and how much it costs in multimodal fusion specifically.

\subsection{In-context learning feature extraction} We follow \citet{Ye2025.Closer} and define the following extraction methods. Let $\mathcal{D}_{tr}$, $\mathcal{D}_{val}$, $\mathcal{D}_{te}$ be the training, validation, and test sets. We define $h(\mathcal{X};\,\mathcal{C}): \mathbb{R}^M \to \mathbb{R}^D$ that maps a set of instances $\mathcal{X}$ to a set of instance encodings in the \emph{query} role given labeled context set $\mathcal{C}$. Analogously, $h(\cdot;\,\mathcal{X})$ maps the set of instances but in the \emph{context} role. Validation and test embeddings are fixed across all schemes as $\phi(\mathcal{X}) = h(\mathcal{X};\,\mathcal{D}_{tr})$ where $\mathcal{X} \in \left\{\mathcal{D}_{val}\cup\mathcal{D}_{te}\right\}$. The three schemes proposed in the prior work differ only in how $\mathcal{D}_{tr}$ embeddings are obtained.

\paragraph{Vanilla ($\star$)} Training instances' embeddings are extracted in the context role:
\begin{equation}\label{eq:van}
  \phi^{\star}(\mathcal{D}_{tr}) = h(\cdot\,;\,\mathcal{D}_{tr})
\end{equation}
\paragraph{Leave-one-fold-out ($\dagger$)} Let's partition $\mathcal{D}_{tr} = F_1 \cup \cdots \cup F_K$ into $K$ disjoint folds. Each fold is encoded as queries against the complement in a such way that every training instance is encoded in the query role exactly once:
\begin{equation}\label{eq:lofo}  
  \phi^{\dagger}_k(F_{k}) = h(F_{k};\,\mathcal{D}_{tr} \setminus F_{k}) \qquad \forall k = 1\dots K
\end{equation}

\paragraph{Non-partitioned ($\wr$)} All training instances are passed simultaneously as both context and queries; their representations are extracted in the query role:
\begin{equation}\label{eq:np}  
  \phi^{\wr}(\mathcal{D}_{tr}) = h(\mathcal{D}_{tr};\,\mathcal{D}_{tr})
\end{equation}

\section{Experiment setup}
\label{sec:exp_design}

Each modality is encoded by its pretrained frozen encoder. This ensures that the unimodal and multimodal baselines receive identical representations as input, so any performance difference between them reflects the downstream module rather than differences in the encoding. For images, we extract the \texttt{CLS} token of a pretrained ViT-B/16~\cite{Dosovitskiy2020.Image}. Since encoders differ in output dimensionality, each modality representation is projected with $\texttt{Linear}(\texttt{LayerNorm}(\cdot))$ to a shared $d \in \{192, 256, 512, 768\}$, keeping parameter counts comparable across encoders. The LayerNorm normalizes each representation to a consistent scale before projection, accounting for differences in output magnitude across encoders. $d$ is a hyperparameter that is selected based on average inner-fold F1. All other hyperparameter details can be found in Appendix~\ref{app:hps}.

\subsection{Models}

\paragraph{Literature baseline} We use TIP~\citep{Du2024.TIP} as a representative advanced multimodal method. It is a carefully designed pipeline: (i) a specialized tabular encoder with a 4-layer Self-Attention~\citep{Vaswani2017.Attention} backbone; (ii) a following fusion module based on 4-layer Cross-Attention~\citep{Vaswani2017.Attention}, with tabular tokens as queries and image tokens as keys and values; (iii) a pretraining stage with a three-component multimodal loss; and (iv) an ensemble trained on representations collected at different stages of the architecture.

\paragraph{Unimodal baseline} As unimodal baselines, we train a linear classifier $\mathbf{W}\mathbf{x} + \mathbf{b}$, where $\mathbf{x} \in \mathbb{R}^d$ is the latent representation of a single modality. Each modality is trained independently and with no access to the other.

\paragraph{Multimodal method} Following \emph{abstract fusion}~\cite{Liang2024a}, we deploy the bilinear product from~\cite{Jayakumar2019} as the fusion module. Given projected tabular embedding $\mathbf{t} \in \mathbb{R}^{d}$, projected image embedding $\mathbf{i} \in \mathbb{R}^{d}$ and learnable parameters $\mathbb{W} \in \mathbb{R}^{d \times C \times d}$, $\mathbf{U},\mathbf{V} \in \mathbb{R}^{d \times C}$, $\mathbf{b} \in \mathbb{R}^{C}$ ($C$ is the number of classes), the output logits are:
\begin{equation}
  \mathbf{t}^\mathsf{T}\mathbb{W}\mathbf{i} + \mathbf{t}^\mathsf{T}\mathbf{U} + \mathbf{V}\mathbf{i} + \mathbf{b}  
\end{equation}
 Setting the output dimension to $C$ makes a clear comparison with the unimodal baseline: if one modality's representation is the zero vector, the bilinear form reduces to a single linear layer, which is exactly the unimodal classifier. Additionally, this fusion subsumes concatenation and tensor fusion.

\subsection{Tabular encoders}
\label{sec:encoders}

For all encoders, representations are extracted from the last layer before the classification head, using default hyperparameters. Encoder selection mostly follows TabArena~\citep{Erickson2025.TabArena} (as of February 26, 2026), covering three classes of tabular encoders.

\paragraph{ICL TFMs}
TabPFNv2~\citep{Hollmann2025.Accurate}, TabDPT~\citep{Ma2025.TabDPT}, and TabICLv2~\citep{Qu2026.TabICLv2} are the primary subjects of study. Each is evaluated under all three extraction schemes described in Section~\ref{sec:icl_shift}, with LOFO($\dagger$) using $K=10$ folds following~\citet{Ye2025.Closer}, yielding $9$ ICL TFM encoder variants in total. Although all three models support ensembling, we fix the ensemble size to one to keep the comparison focused on representation quality.

\paragraph{Non-ICL TFM}
TARTE~\citep{Kim2025} belongs to the non-ICL family of tabular foundation models (other examples are XTab~\citep{ZhuXtab2023}, TransTab~\citep{WangTransTab2022}, and UniTabE~\citep{YangUniTabE2023}). It has no context/query asymmetry, making it a natural reference point.

\paragraph{Baselines}
TabM~\citep{Gorishniy2024.TabM} is pretrained from scratch directly on the target task using only the tabular modality, serving as a baseline for tabular deep learning. TabM-SSL shares the same architecture and hyperparameters but is pretrained on reconstruction and masking objectives~\citep{Rubachev2022.Revisiting}, serving as a lower bound for SSL methods. Unlike ICL TFMs, TabM suffer a sharp performance drop when the ensemble size is reduced to one. We therefore retain their full ensembles and aggregate member embeddings via a learned weighted average $w \in \mathbb{R}^N$, where $N$ is the number of members. Finally, similarly to~\citep{Kim2025}, we include Tabular Vectorizer (TabVec) from skrub~\citep{Skrub}, a non-learned preprocessing baseline that serves as an overall lower bound.

\subsection{Datasets}
Following~\citet{Tang2024a}, we evaluate on seven tabular-image classification datasets previously spanning medical imaging, art, vehicle, and pet adoption domains: DVM~\citep{Huang2022a}, Petfinder~\citep{PetFindermyAdoptionPredictionPetFindermyAdoptionPredictiona}, Wikiart~\citep{wikiart2025}, HAM10000~\citep{Codella2019.Skin, SkinCancerMNISTSkinCancerMNISTHAM10000}, CCD~\citep{CarCrashDatasetCarCrashDatasetCCD}, COVID~\citep{Cohen2020,COVIDGitHub}, and Artm~\citep{ArtPriceDatasetArtPriceDataset}. Each dataset pairs a tabular feature table with at least one image per instance. Instances with a completely missing modality are omitted. Table~\ref{tab:datasets} provides an overview. Preprocessing details are available in the code.

\begin{table}[h]
\centering
\footnotesize
\caption{Datasets overview.
  \textbf{N}: number of instances.
  \textbf{\#Cat.}/\textbf{\#Num.}: number of categorical/continuous features.
  \textbf{\#Cls.}: number of target classes.
  The last four columns report quantiles at [2.5, 25, 50, 75, 97.5]\%.
  \textbf{NA/row}: number of missing values per row.
  \textbf{NA/col}: number of missing values per column.
  \textbf{Imgs/inst}: number of images per instance.
  \textbf{Img $\sqrt{W{\times}H}$}: square root of image area.}
\label{tab:datasets}

\setlength{\tabcolsep}{3.5pt}
\begin{tabular}{lcccccccc}
\toprule
\textbf{Dataset} & \textbf{N} & \textbf{\#Cat.} & \textbf{\#Num.} & \textbf{\#Cls.} & \textbf{NA/row} & \textbf{NA/col} & \textbf{Imgs/inst} & \textbf{Img $\sqrt{W{\times}H}$} \\
\midrule
DVM       & 34{,}272 & 6 & 15 & 10 & 0/0/0/1/8 & 0/10/294/2376/5425 & 1/4/6/7/16 & 300/300/300/300/300 \\
Petfinder& 14{,}652 & 12 & 7 & 5 & 0/1/2/2/3 & 0/0/0/0/10455 & 1/2/3/5/13 & 270/346/346/416/612 \\
Wikiart   & 11{,}335 & 2 & 0 & 10 & 0/0/0/0/0 & 0/0/0/0/0 & 1/1/1/1/1 & 1421/1538/1588/1679/1954 \\
HAM       & 7{,}470 & 2 & 1 & 7 & 0/0/0/0/0 & 0/0/0/26/49 & 1/1/1/2/3 & 520/520/520/520/520 \\
CCD       & 1{,}500 & 2 & 0 & 2 & 0/0/0/0/0 & 0/0/0/0/0 & 1/1/1/1/1 & 960/960/960/960/960 \\
COVID     & 845 & 7 & 7 & 2 & 4/7/8/9/11 & 0/98/595/798/826 & 1/1/1/1/1 & 323/664/954/1644/2891 \\
Artm      & 635 & 1 & 2 & 7 & 0/0/0/0/1 & 0/0/0/31/58 & 1/1/1/1/1 & 518/648/653/653/653 \\
\bottomrule
\end{tabular}
\end{table}

\subsection{Metrics}
\label{sec:metric}
\paragraph{RPR} The same absolute F1 difference carries different weight across datasets, depending on how tightly methods cluster near the performance ceiling. Relative Percentile Rank (RPR) addresses this by measuring each method's performance relative to a common reference, the image-only baseline, using per-dataset percentile ranks. RPR is computed per dataset. For each of the $n$ tabular encoders, there is one multimodal model and one tabular-only unimodal baseline, giving $2n$ methods in total. These $2n$ methods together with the image-only baseline are ranked by F1 to percentile ranks (1 = best, 0 = worst). The image-only baseline's rank is then subtracted from the $2n$ remaining methods' ranks.  As a result, a positive RPR means the method outperforms the image-only baseline in relative rank, and a negative value means it does not.

\paragraph{EDS} Motivated by the representation shift identified in Section~\ref{sec:icl_shift}, we quantify it with Maximum Mean Discrepancy (MMD)~\citep{Gretton2012} using a radial basis function kernel averaged over log-spaced bandwidths $\bigl\{ e^{-2 + \frac{6k}{5}}: k\in\left\{0\dots 5\right\} \bigr\}$. To obtain a gap measure relative to data/encoder-intrinsic variation, we subtract $\mathrm{MMD}\bigl(\phi(\mathcal{D}_{val}), \phi(\mathcal{D}_{te})\bigr)$ and call the result Excess Distribution Shift (EDS):
\begin{equation}\label{eq:eds}
  \begin{array}{l}
   \text{EDS}(\mathcal{D}_{tr}, \mathcal{D}_{val}) = \mathrm{MMD}\bigl(\phi(\mathcal{D}_{tr}), \phi(\mathcal{D}_{val})\bigr) - \mathrm{MMD}\bigl(\phi(\mathcal{D}_{val}), \phi(\mathcal{D}_{te})\bigr)\\[3pt]
   \text{EDS}(\mathcal{D}_{tr}, \mathcal{D}_{te}) = \mathrm{MMD}\bigl(\phi(\mathcal{D}_{tr}), \phi(\mathcal{D}_{te})\bigr) - \mathrm{MMD}\bigl(\phi(\mathcal{D}_{val}), \phi(\mathcal{D}_{te})\bigr)
  \end{array}
\end{equation}
As the result, a positive value means train embeddings are further from the test distribution than two ordinary held-out splits are from each other. 

\section{Experiment results}\label{sec:results}

\begin{figure}[h]
  \centering
  \includegraphics[width=\linewidth]{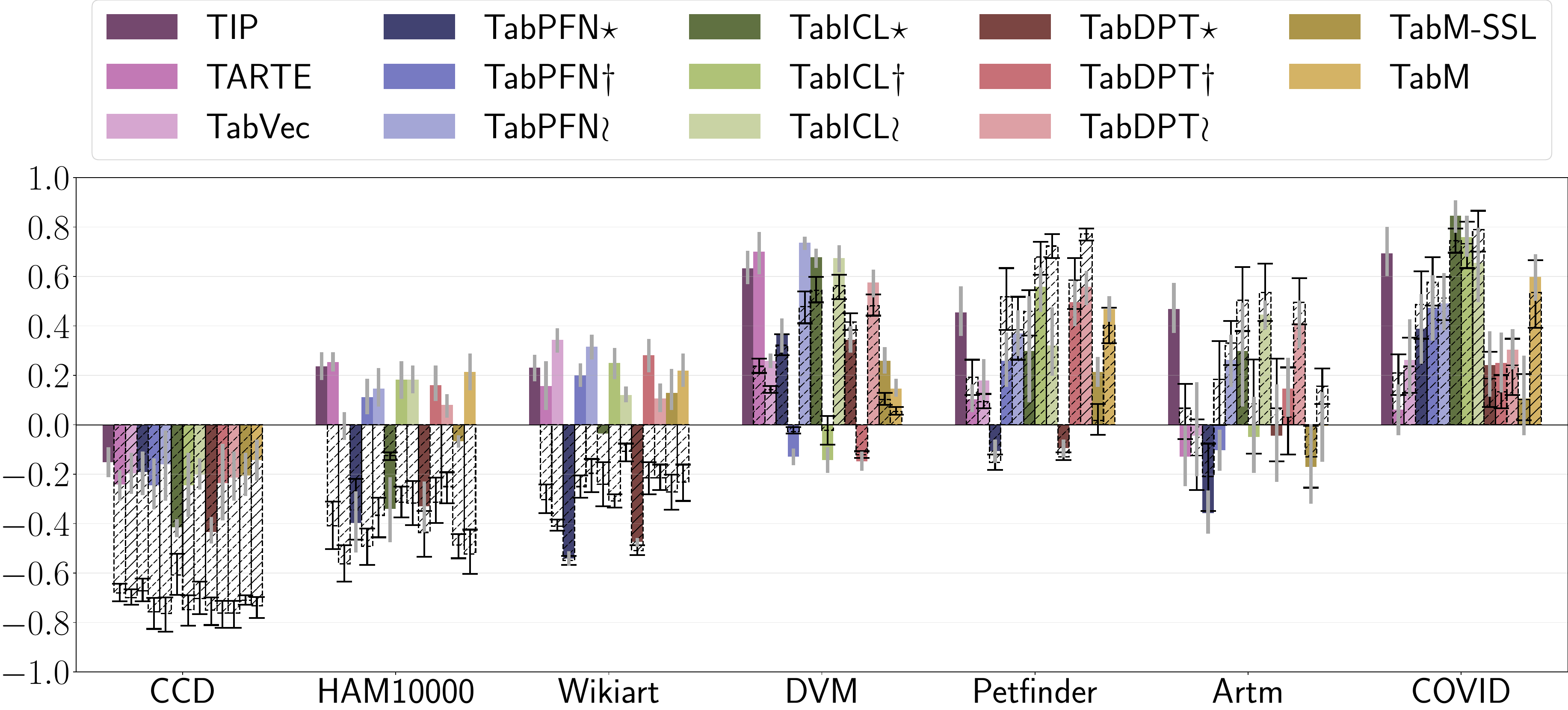}
  \caption{\textbf{RPR $\uparrow$.} See the section~\ref{sec:metric} for details. Filled bars show multimodal performance (color indicates tabular encoder). Hatched bars show the corresponding unimodal tabular performance on the same tabular encoder. Error bars (uncapped gray for multimodal, capped black for unimodal tabular) indicate 95\% confidence intervals. ICL features extractions are $\star=$ Vanilla, $\dagger=$ LOFO, and $\wr=$ NP.}
  \label{fig:rpr}
\end{figure}

We begin with the overall picture. Figure~\ref{fig:rpr} requires careful reading. For each tabular encoder, the multimodal bar (filled) and its unimodal counterpart (hatched) are overlaid on each other, allowing direct visual comparison. TIP is the only exception, as it is an end-to-end multimodal method that cannot be reduced to a unimodal baseline by design. Error bars are 95\% confidence intervals: capped black for unimodal, uncapped gray for multimodal. Overlapping intervals indicate no statistically significant difference. All values are measured in RPR relative to the image-only baseline, meaning a bar above zero indicates the model outperforms image alone. With this in mind, we can turn to the analysis.

\subsection{Evaluation on a sole encoder}
\label{sec:sole_encoder_story}

To assess how encoder choice affects method comparisons, we evaluate TIP~\citep{Ye2025.Closer}, a carefully engineered multimodal method, against our unimodal and multimodal baselines while varying only the tabular embedding (Figure~\ref{fig:rpr}). Scores are reported as (W/T/L). With TabVec, TIP holds a clear advantage in the unimodal setting (7/0/0) and a positive edge in the multimodal setting (3/3/1). Switching to the stronger TabM encoder, TIP continues to lead in the unimodal setting (5/2/0) while the multimodal margin narrows to (2/5/0). ICL encoders shift the picture further. Against TabPFN-NP($\wr$), the scores are (5/2/0) in the unimodal setting and (1/5/1) in the multimodal setting. Against TabICL-NP($\wr$), they are (3/3/1) and (0/7/0). These results highlight a general sensitivity of multimodal comparisons to the choice of tabular encoder, and conclusions drawn from a sole encoder may not hold across the full encoder landscape.

 




\subsection{Likely multimodal datasets} 

 Figure~\ref{fig:rpr} reveals different dataset types. We call CCD \textit{image-dominated}: all bars, unimodal and multimodal alike, fall below zero, indicating that the tabular embedding carries no predictive signal and in fact hurts performance. At the other extreme, we call Petfinder, ArtM, and COVID \textit{tabular-dominated}: unimodal performance matches or even exceeds multimodal. We call HAM10000, WikiArt, and DVM \textit{likely multimodal}, where fusion significantly outperforms the unimodal baseline across most tabular encoders. Within this group, DVM is tabular-centric, while HAM10000 and WikiArt are image-centric. For a detailed breakdown of modality contributions for \textit{likely-multimodal} datasets, see Appendix~\ref{sec:modality_contribution}.

Remarkably, in modality-dominated datasets, fusion can degrade performance below the unimodal baseline. Intuitively, a fusion module would learn to suppress the uninformative modality and recover the unimodal solution. In practice this does not happen: in CCD (image-dominated), adding tabular representations hurts, and in Petfinder (tabular-dominated), adding image representations hurts. Notably, TIP exhibits the same degradation, suggesting the problem is not specific to the bilinear fusion design. Whether it extends to other fusion architectures remains an open question and a natural direction for future work.
 \begin{figure}[h]
  \centering
  \includegraphics[width=0.8\linewidth]{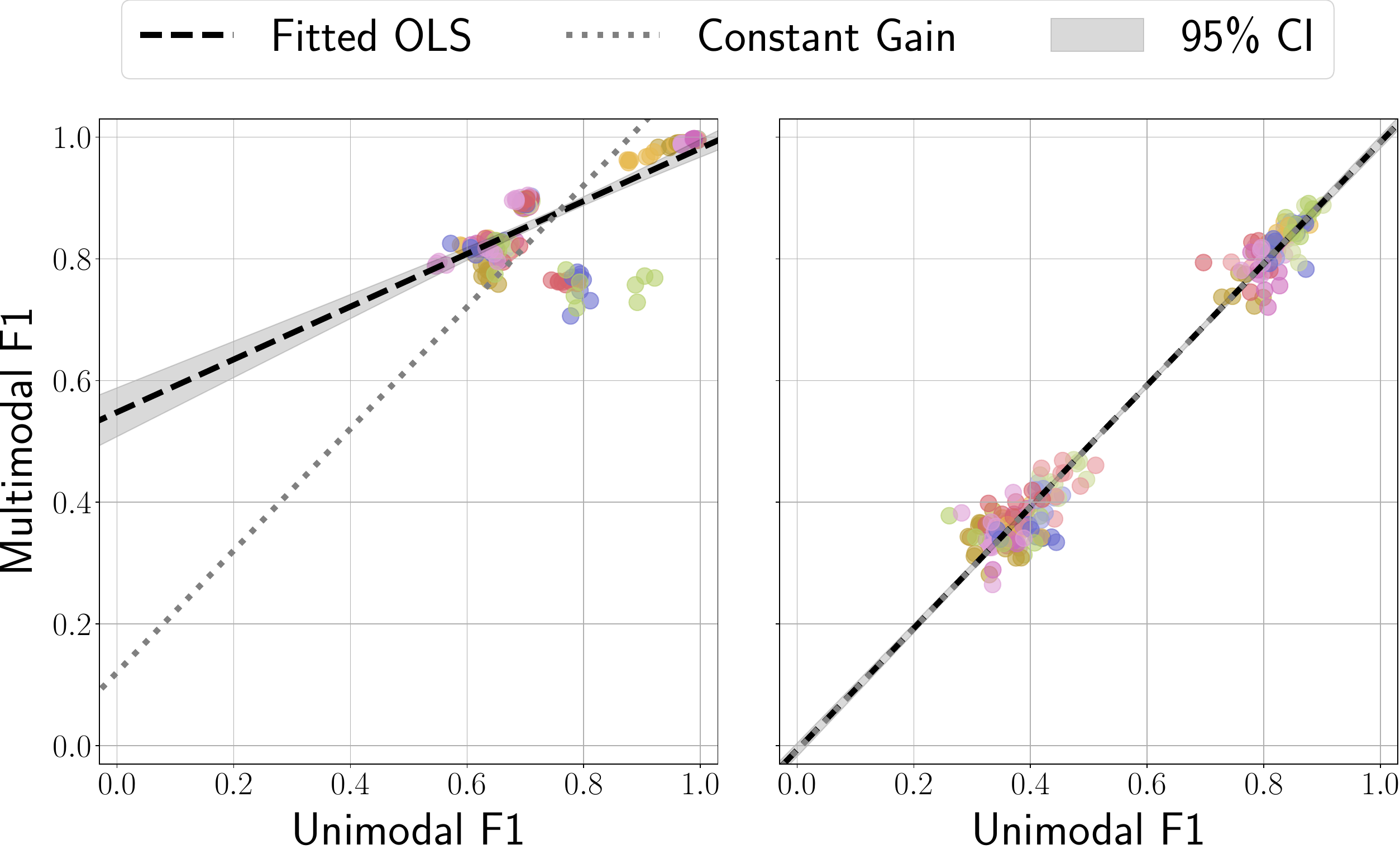}
  \caption{Multimodal F1 against Unimodal F1.~\textbf{Left}: \emph{Likely} multimodal datasets. \textbf{Right}: \emph{Tabular-dominated} multimodal datasets. The dashed line shows the OLS fit with its 95\% confidence band (reflecting uncertainty in the estimated slope and intercept), the dotted line shows constant gain with a slope of 1, encoder-independent lift. Colors are tabular embeddings following the same coding as on Figure~\ref{fig:rpr}.}
  \label{fig:mm_vs_uni}
\end{figure}

\subsection{Multimodal lift is encoder-dependent} We analyze whether multimodal fusion $(Y)$ reliably lifts performance above the unimodal baseline $(X)$, and whether this lift $(Y-X)$ depends on encoder quality. To this end, we fit an ordinary least squares (OLS) regression $Y = \beta X + b$ and subtract $X$ from both sides, obtaining $Y - X = (\beta - 1)X + b$. in such formulation, the null hypothesis is $H_0\colon \beta = 1$. Under the null hypothesis, the lift is a constant $b$ (if it exists, $b\gg0$), fully independent of encoder quality. A slope below $1$ indicates that the fusion advantage shrinks as the unimodal baseline improves, suggesting that part of the gain compensates for encoder weakness rather than capturing genuine multimodal interactions. With a strong encoder, the residual lift is more likely to reflect a real advantage of combining modalities.


Figure~\ref{fig:mm_vs_uni} shows the results for each dataset group. For \emph{tabular-dominated} datasets, $\hat{\beta} = 0.999$ ($95\%$ CI:\ $[0.979, 1.018]$, one-sided $p = 0.442$) with intercept $\hat{b} = -0.007$. The slope is close to $1$ and the intercept near zero, consistent with multimodal and unimodal performance being approximately equal in this group. For \emph{likely} multimodal datasets, $\hat{\beta} = 0.434$ ($95\%$ CI:\ $[0.381, 0.486]$, one-sided $p = 2.5 \times 10^{-57}$) with intercept $\hat{b} = 0.548$. Here the slope is below $1$, suggesting that encoder quality may shape the multimodal lift. The larger intercept points to a base contribution from the image modality. The complement $(1 - \hat{\beta}) \approx 0.566$ offers a more precise reading: within the observed range, a one-unit increase in unimodal F1 is associated with a reduction of approximately $0.566$ in the multimodal advantage. This is consistent with stronger tabular encoders closing the gap with fusion, though the estimate assumes linearity and should be read as an observational, not causal, summary.

\subsection{Naive baseline}

\begin{figure}[h]
  \centering
   \begin{subfigure}[c]{0.49\textwidth}
    \centering
      \includegraphics[width=\linewidth]{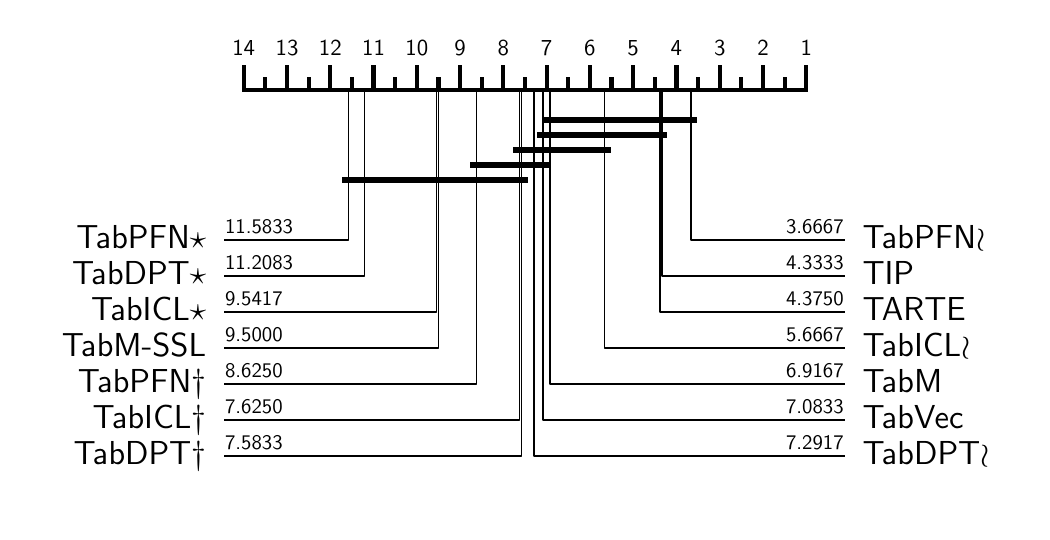}
      \subcaption{\textit{Likely} Multimodal}
  \end{subfigure}
  \hfill
  \begin{subfigure}[c]{0.49\textwidth}
    \centering
    \includegraphics[width=\linewidth]{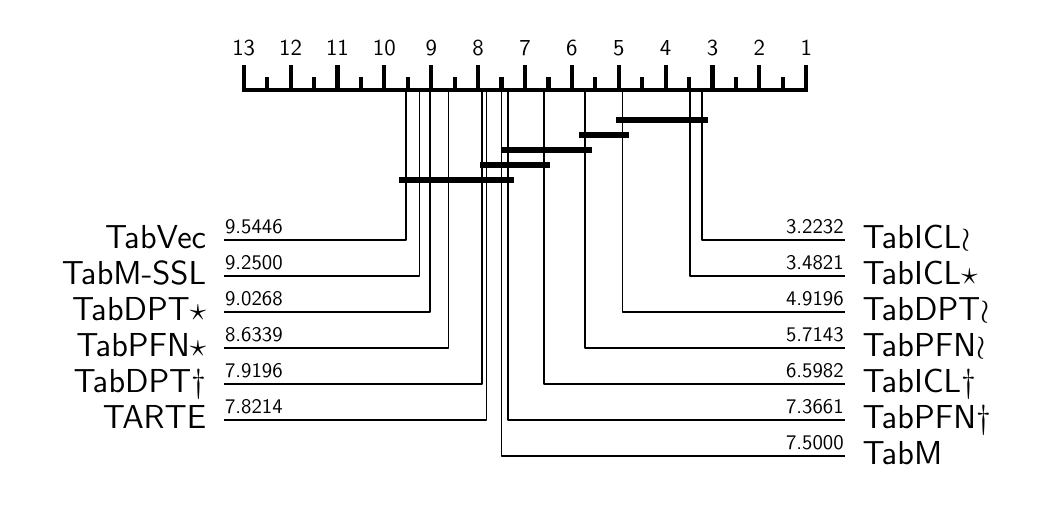}
    \subcaption{Tabular Unimodal Baseline}
  \end{subfigure}
  \caption{Critical Difference Diagrams on F1, where horizontal bar indicates the absence of statistical significance. \textbf{Left}: Multimodal comparison across \emph{likely} multimodal datasets.   Critical Difference Diagram over all multimodal datasets can be viewed on Figure~\ref{fig:cdd_mm}.\textbf{Right}: Tabular unimodal baseline comparison across all datasets.}
  \label{fig:cdd}
\end{figure}

Section~\ref{sec:sole_encoder_story} compared TIP and the naive baseline by win-rate. Figure~\ref{fig:cdd} examines whether those differences reach statistical significance. In the \emph{likely} multimodal setting, a bilinear fusion model paired with a strong encoder (e.g.\ TabPFN$\wr$) is competitive with TIP~\citep{Ye2025.Closer}. TIP is a carefully engineered method with a median of 7.5M trainable parameters (mean 10.4M) and an additional pretraining phase. This corresponds to on average \textbf{13.2$\times$} more parameters than the best-performing baseline (95\% CI: 1.5$\times$-29.9$\times$).

Turning to extraction schemes, NP($\wr$) outperforms LOFO($\dagger$) in both the unimodal and multimodal settings, contrary to the recommendation of~\citet{Ye2025.Closer}. We suspect that this may be related to the smaller context available to LOFO($\dagger$) relative to NP($\wr$). Our results also support the concern raised by~\citet{Ye2025.Closer} regarding Vanilla($\star$): in the multimodal setting, it degrades performance across all three ICL models. In the unimodal setting, the degradation is statistically significant for TabPFN and TabDPT but not for TabICL. Section~\ref{sec:icl_repr_shift_comparison} examines these extraction schemes in more detail.

The diagram also shows that TabPFN$\wr$ is the only ICL encoding to surpass TabVec, a non-pretrained baseline. TARTE matches the best-performing tabular embedding despite receiving no labels during encoding. 

\subsection{Caveat of in-context learning}

The consistently lower RPR of Vanilla($\star$) encodings across all ICL TFMs in Figure~\ref{fig:rpr} suggests that the representation shift is not limited to TabPFNv2.

\begin{figure}[h]
\centering
\begin{minipage}[c]{0.537\textwidth}
  \centering
  \vspace*{0.825em}
  \includegraphics[width=\linewidth]{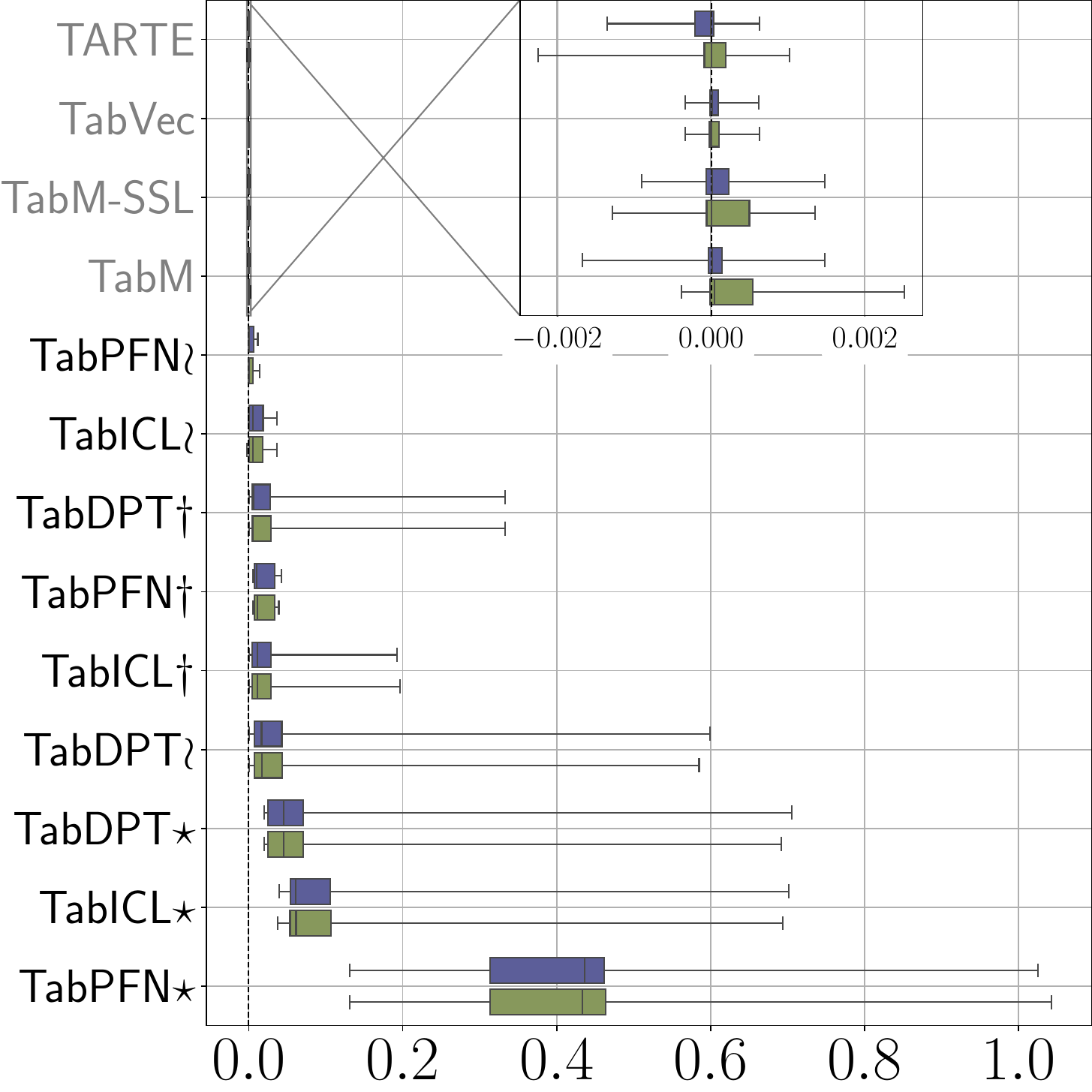}
\end{minipage}%
\hfill%
\begin{minipage}[c]{0.45\textwidth}
  \centering
  {\captionsetup[subfigure]{labelformat=empty, skip=0pt, position=top}
  \centering
\setlength{\tabcolsep}{2pt}
\renewcommand{\arraystretch}{2}

\begin{tabular}{ccc}
\subcaptionbox{F1: 0.8267}{\includegraphics[width=0.32\textwidth]{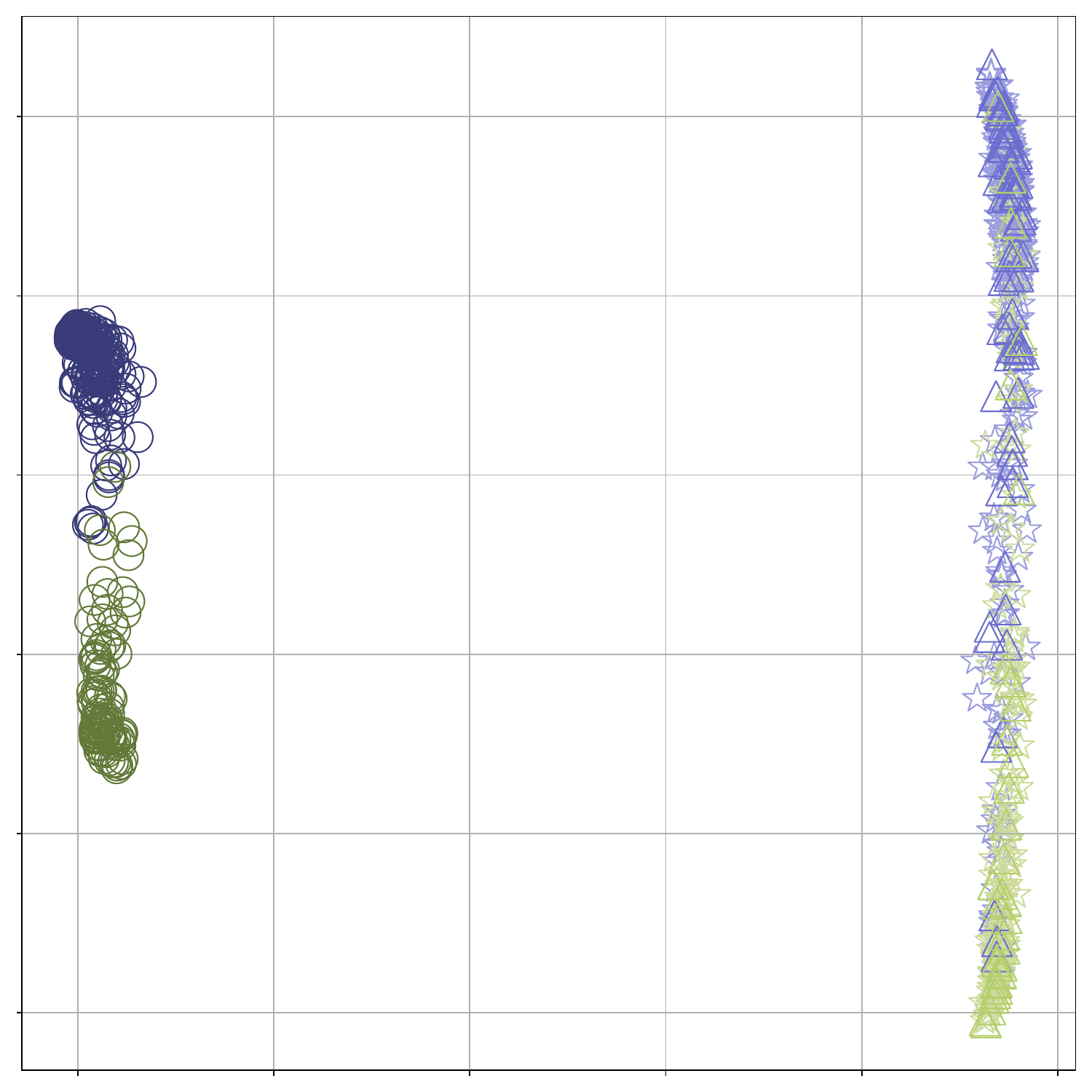}} &
\subcaptionbox{F1: 0.8466}{\includegraphics[width=0.32\textwidth]{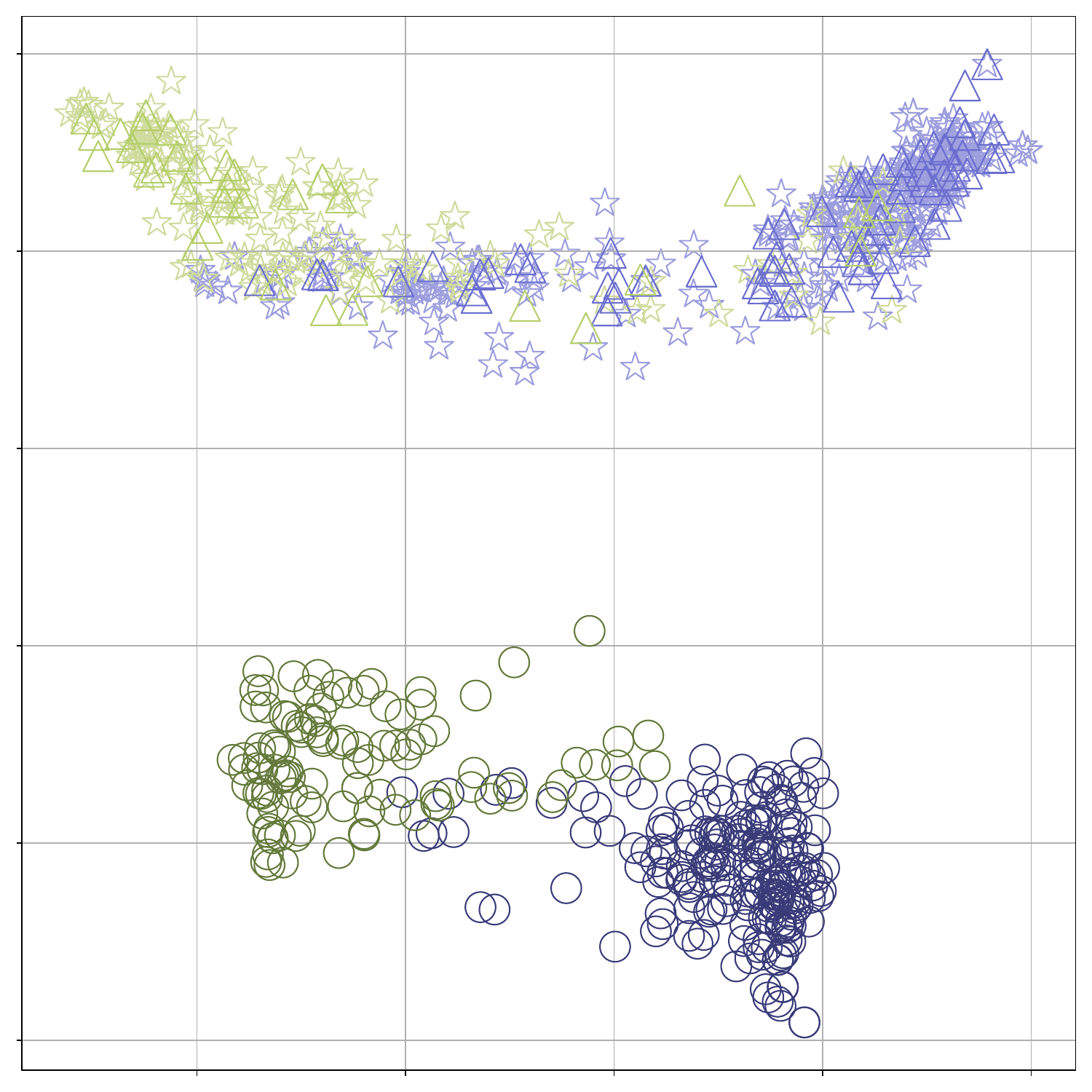}} &
\subcaptionbox{F1: 0.8019}{\includegraphics[width=0.32\textwidth]{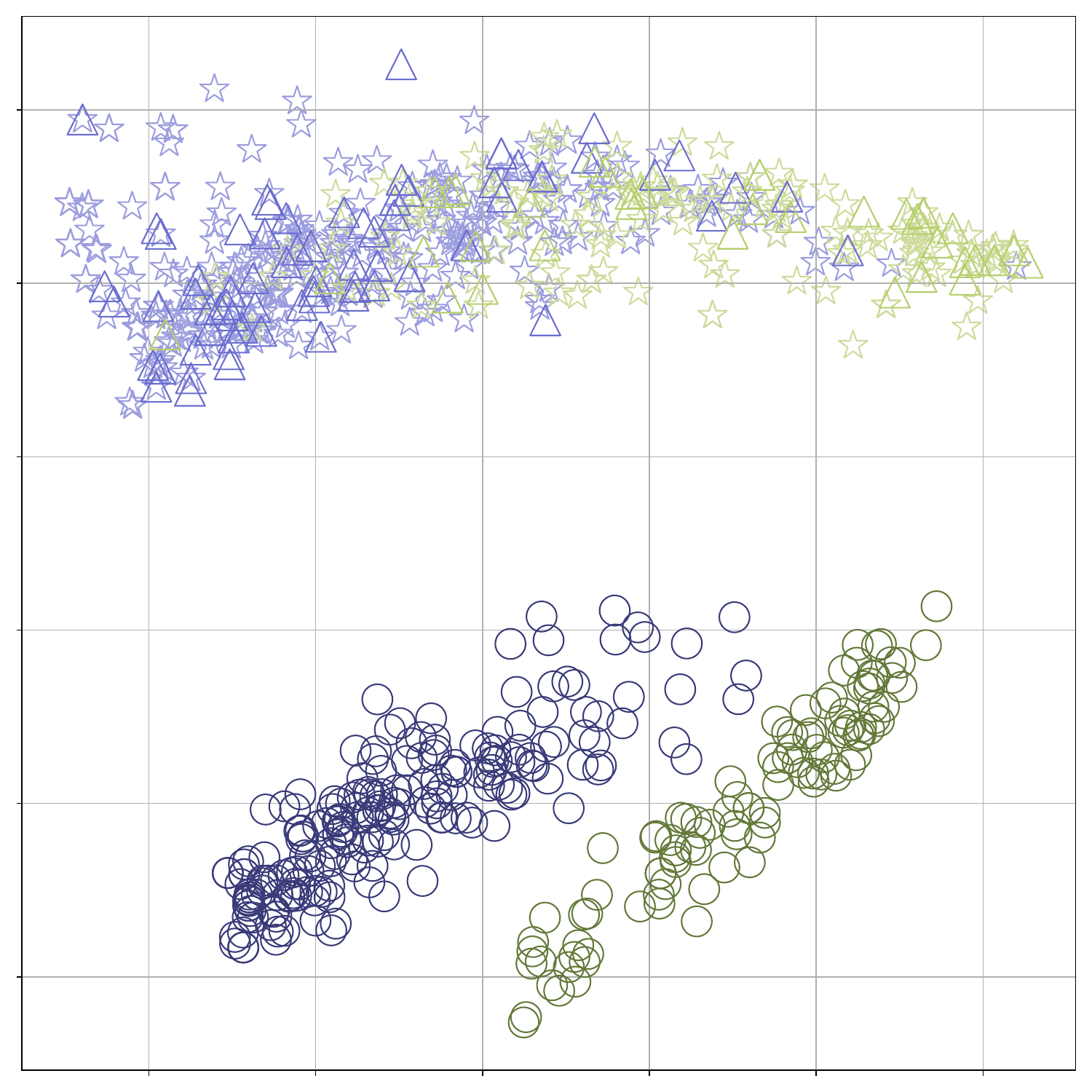}} \\

\subcaptionbox{F1: 0.8165}{\includegraphics[width=0.32\textwidth]{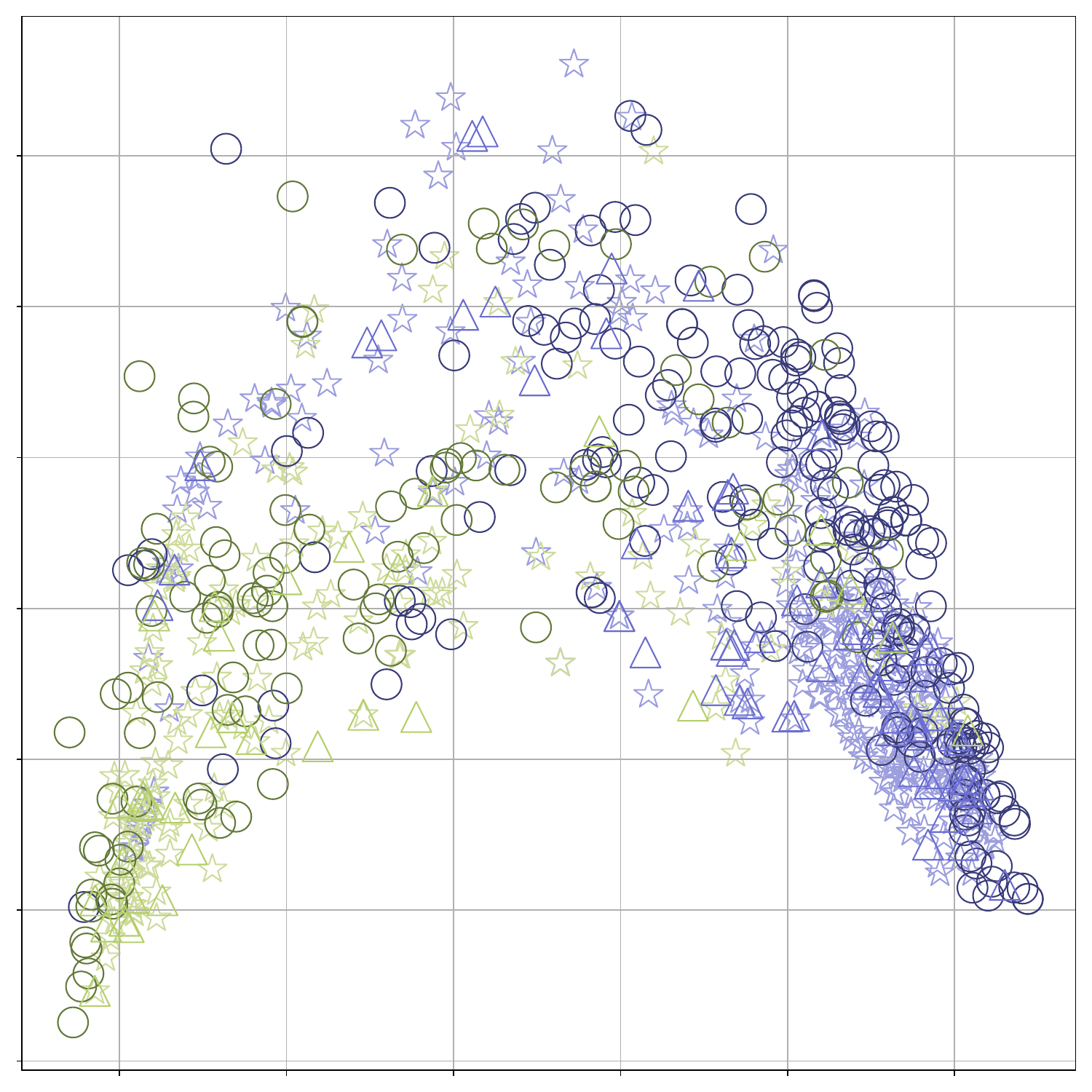}} &
\subcaptionbox{F1: 0.8498}{\includegraphics[width=0.32\textwidth]{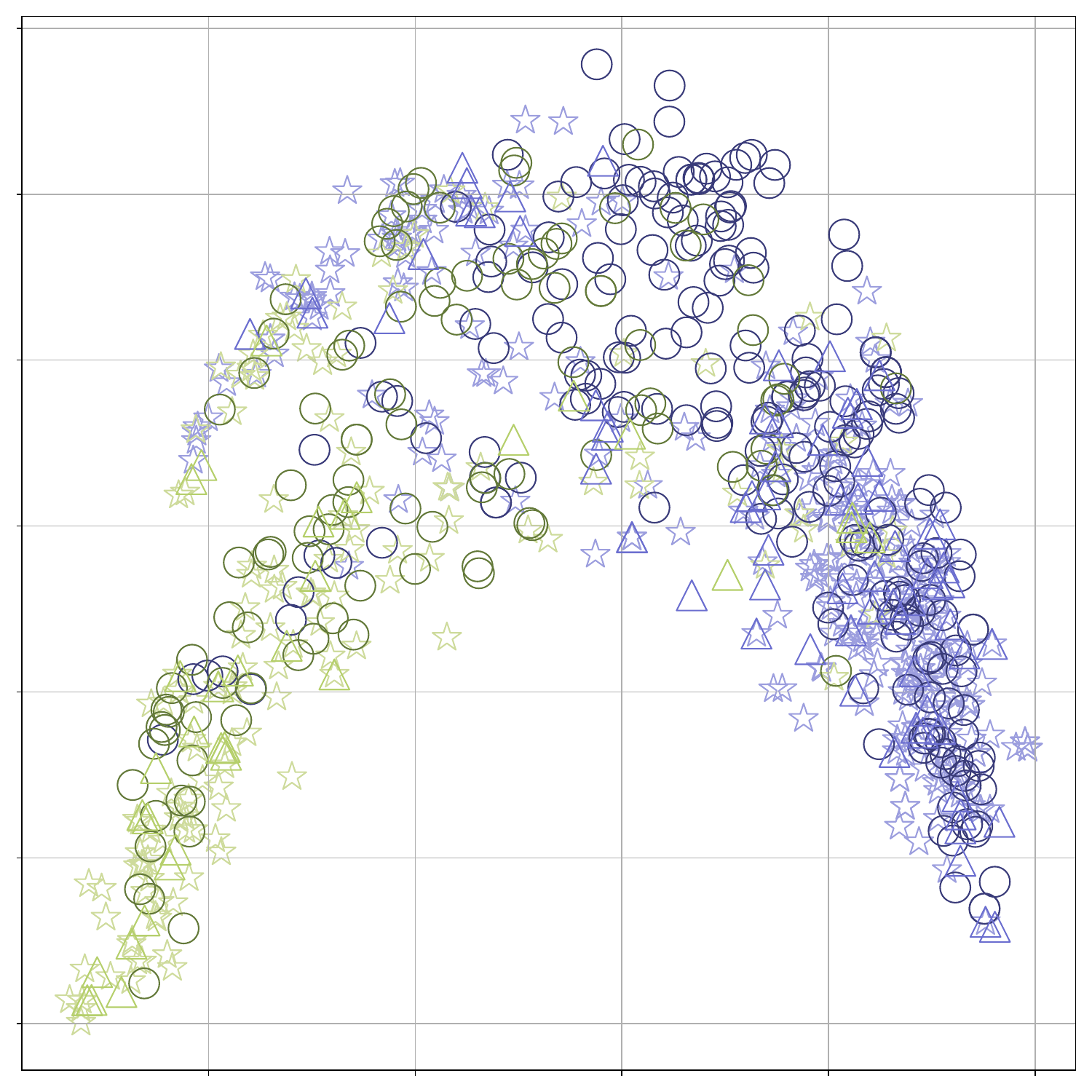}} &
\subcaptionbox{F1: 0.7929}{\includegraphics[width=0.32\textwidth]{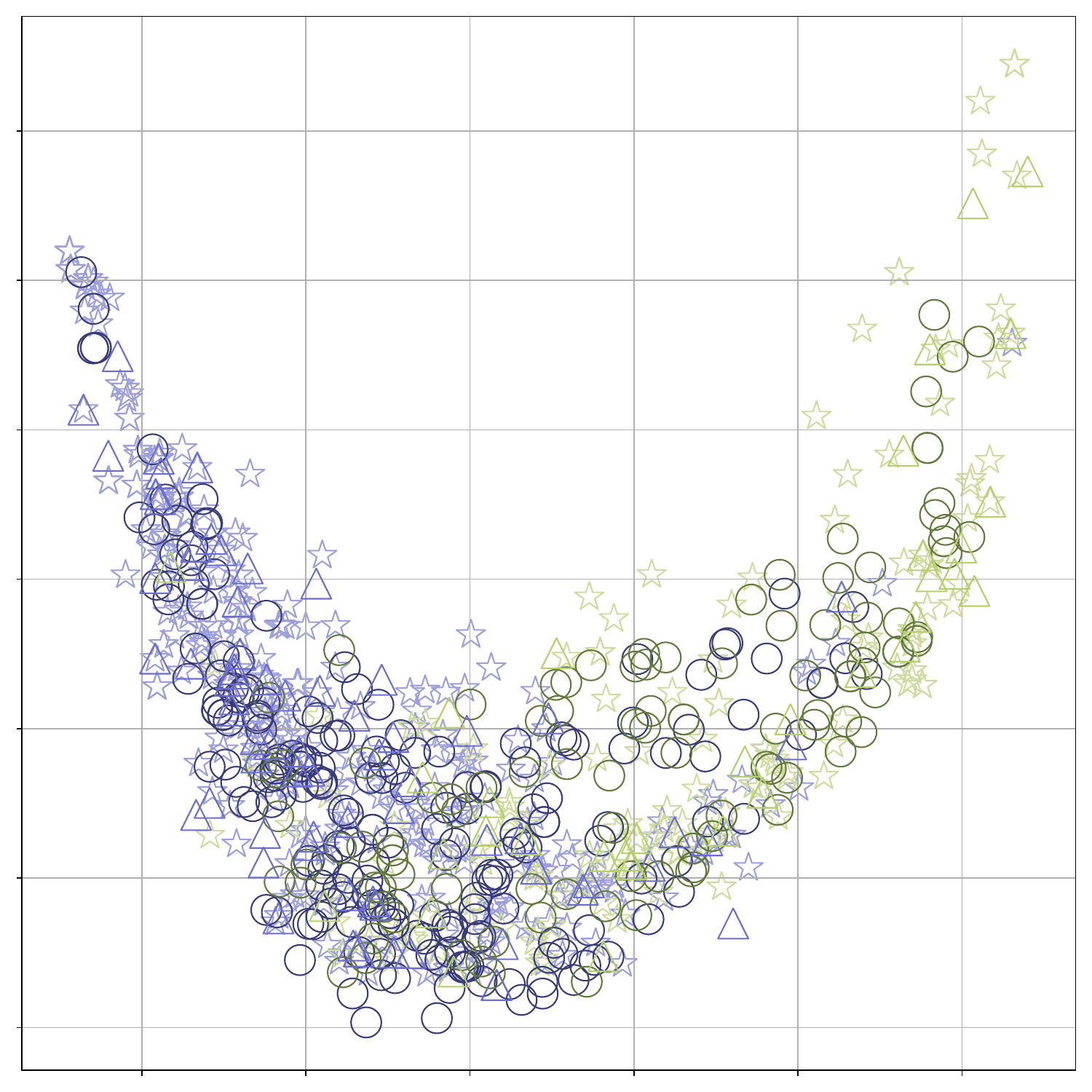}} \\

\subcaptionbox{F1: 0.8215}{\includegraphics[width=0.32\textwidth]{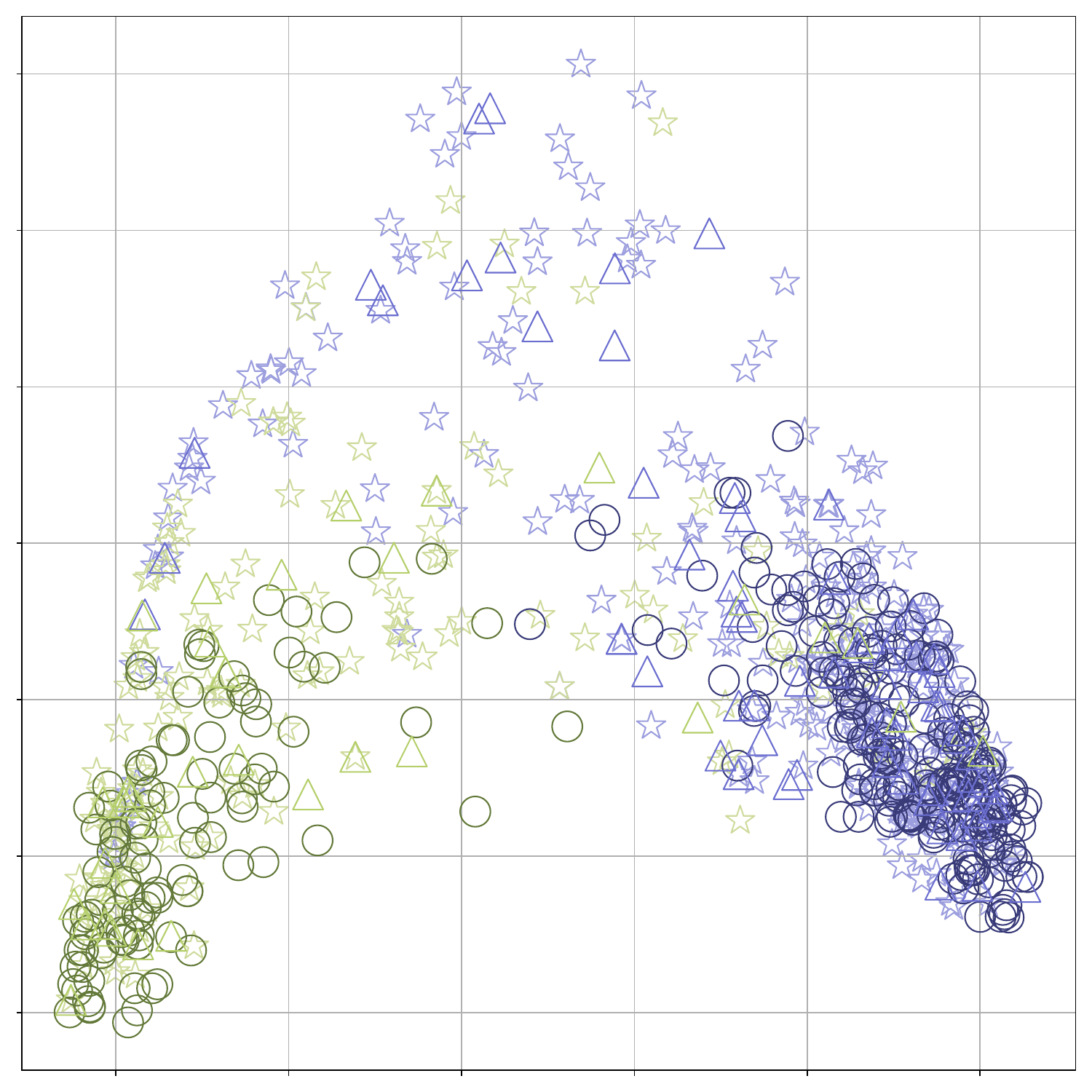}} &
\subcaptionbox{F1: 0.8547}{\includegraphics[width=0.32\textwidth]{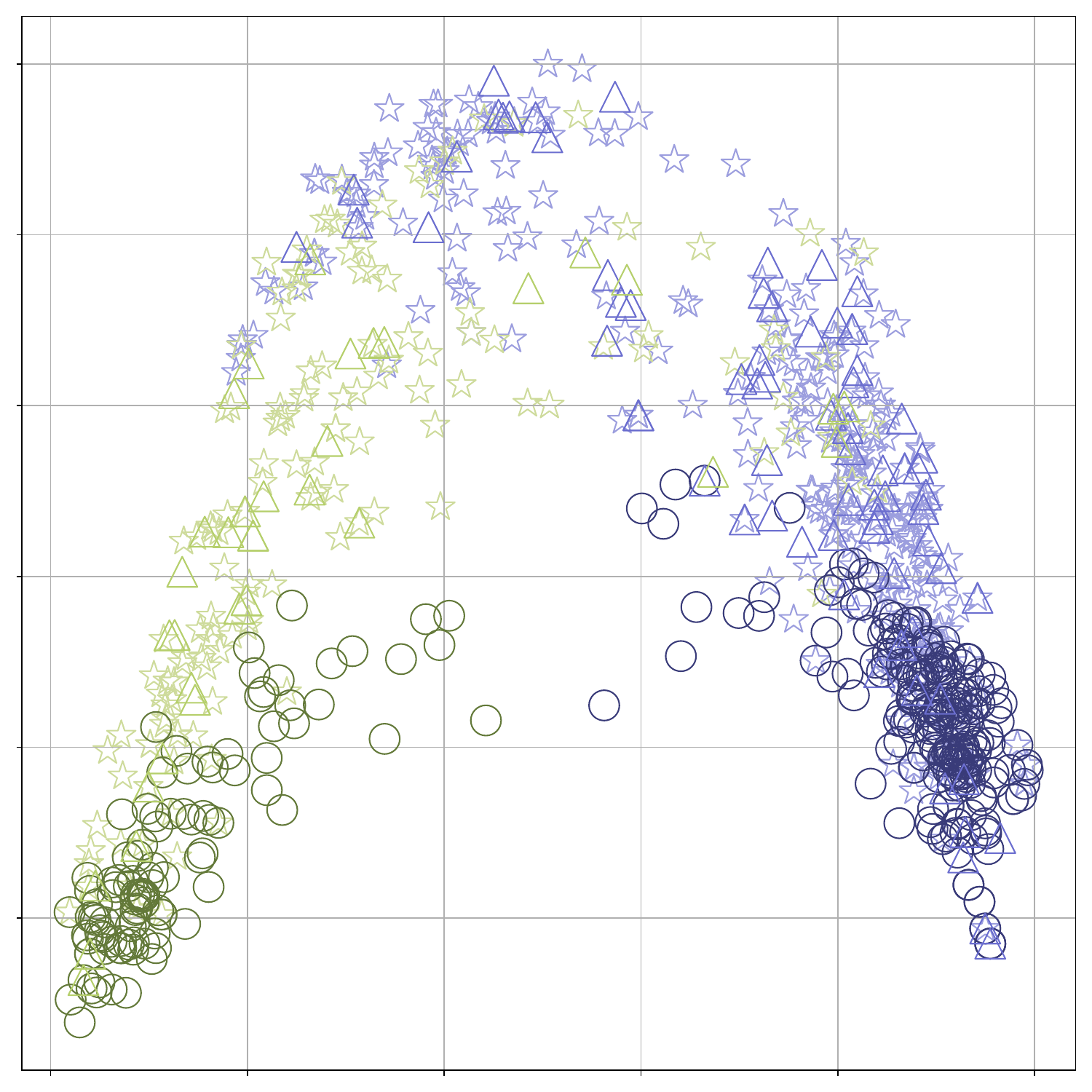}} &
\subcaptionbox{F1: 0.8160}{\includegraphics[width=0.32\textwidth]{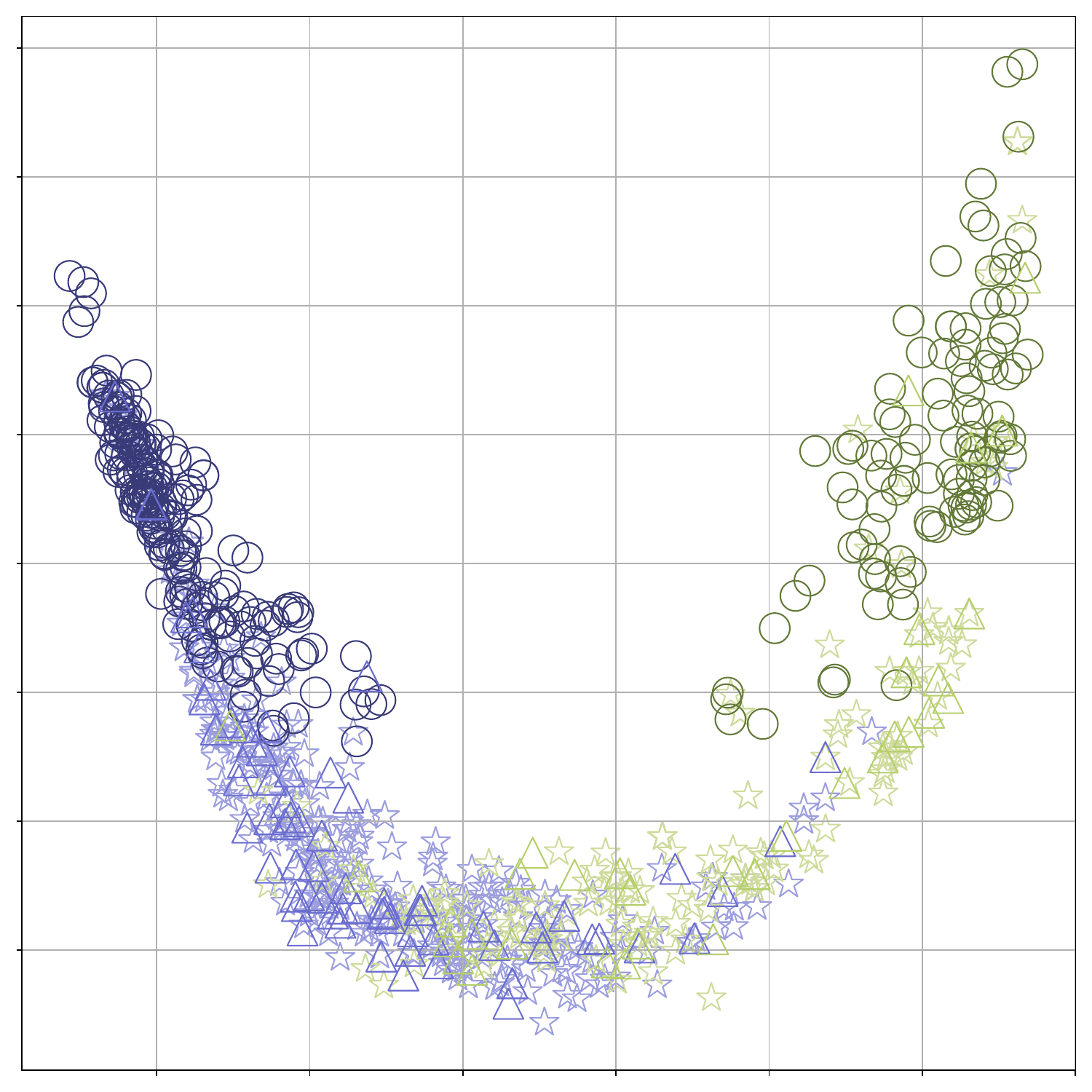}}\\[-3pt]
\textbf{(a) TabPFN} & \textbf{(b) TabICL} & \textbf{(c) TabDPT} \\
\end{tabular}
}
\end{minipage}
\caption{\textbf{Left:}  $\text{EDS}(\mathcal{D}_{tr}, \mathcal{D}_{val})$ in blue and  $ \text{EDS}(\mathcal{D}_{tr}, \mathcal{D}_{te})$ in green (Eq.~\ref{eq:eds}). Boxplots are  aggregated over datasets over each encoder-scheme pair. Positive values exceed data-intrinsic variation.
\textbf{Right:} 2D PCA on COVID (rows top to bottom: Vanilla($\star$), LOFO($\dagger$), NP($\wr$); columns left to right: TabPFN, TabICL, TabDPT). Dark circles and light triangles/stars are train and val/test respectively. Color reflects class.}
\label{fig:shift_grid}
\end{figure}

Figure~\ref{fig:shift_grid} (right) shows PCA projections of train and test embeddings for all three models on the COVID dataset, as an example. Rows correspond to feature extraction schemes (top to bottom: Vanilla $\star$, LOFO $\dagger$, NP $\wr$) and columns to models (left to right: TabPFN, TabICL, TabDPT). Under Vanilla encoding, context and query embeddings form visibly separated clusters in every model. The shift is not only visible in TabPFNv2 but also in TabDPT and TabICLv2. LOFO ($\dagger$) and NP $\wr$ reduce the separation in all three cases. Subplot titles report the F1 score of a tuned unimodal baseline on the validation split (see Section~\ref{sec:exp_design}). 

While the PCA projections give an intuitive picture, they are limited to a single dataset and cannot support a general conclusion. Therefore, we quantify the shift across all available datasets with $\text{EDS}$ (see Eq.~\ref{eq:eds}). Figure~\ref{fig:shift_grid} (left) shows $\text{EDS}(\mathcal{D}_{tr}, \mathcal{D}_{val})$ in blue and $\text{EDS}(\mathcal{D}_{tr}, \mathcal{D}_{te})$ in green as boxplots per encoder, where each point corresponds to one dataset. Non-ICL encoders (TARTE, TabM and TabM-SSL), as expected, produce near-zero gaps by construction. All three ICL TFMs show positive gaps under Vanilla ($\star$) across datasets, though the magnitude differs: TabPFNv2 produces the largest separation, followed by TabICLv2 and TabDPT. LOFO($\dagger$) and NP($\wr$) reduce both metrics consistently across all ICL TFMs. These results suggest that the shift persists in the original latent space and is not specific to TabPFNv2. Both extraction schemes help mitigate it consistently across models. Whether the shift translates into a performance difference is examined next.
\label{sec:icl_repr_shift_comparison}

\begin{table}[h]
\setlength{\tabcolsep}{5pt}
\centering
\caption{Spearman correlation between $\text{EDS}(\mathcal{D}_{tr}, \mathcal{D}_{te})$ and F1 Rank for each ICL TFM. Tabular-dominated datasets (Tab.\ Dom.) have a redundant image modality, approximating the unimodal setting. Likely multimodal datasets yield a performance benefit from combining both modalities.}
\begin{tabular}{lccccccc}
\toprule
&& \multicolumn{2}{c}{TabPFN} & \multicolumn{2}{c}{TabICL} & \multicolumn{2}{c}{TabDPT} \\
\cmidrule(lr){3-4} \cmidrule(lr){5-6} \cmidrule(lr){7-8}
Datasets&Model& $\rho$ & $p$ & $\rho$ & $p$ & $\rho$ & $p$ \\
\midrule
Tab.Dom. & Uni & $-0.017$ & $9.1 \times 10^{-1}$ & $0.292$ & $\mathbf{4.4 \times 10^{-2}}$ & $0.052$ & $7.3 \times 10^{-1}$ \\
Tab.Dom.& MM & $-0.447$ & $\mathbf{1.5 \times 10^{-3}}$ & $-0.205$ & $1.6 \times 10^{-1}$ & $-0.118$ & $4.2 \times 10^{-1}$ \\
Likely & Uni & $-0.577$ & $\mathbf{1.8 \times 10^{-5}}$ & $-0.438$ & $\mathbf{1.9 \times 10^{-3}}$ & $-0.226$ & $1.2 \times 10^{-1}$ \\
Likely & MM & $-0.727$ & $\mathbf{4.9 \times 10^{-9}}$ & $-0.361$ & $\mathbf{1.2 \times 10^{-2}}$ & $-0.366$ & $\mathbf{1.1 \times 10^{-2}}$ \\
\bottomrule
\end{tabular}
\label{tab:mmd_corr}
\end{table}

Table~\ref{tab:mmd_corr} reports Spearman correlations between $\text{EDS}(\mathcal{D}_{tr}, \mathcal{D}_{te})$ and F1 rank. TabPFNv2 shows a significant negative correlation in three of four conditions, absent only in the \emph{tabular-dominated} unimodal setting. The effect is larger in \emph{likely} multimodal datasets. Within that group, the multimodal baseline shows a stronger effect than the unimodal one ($\rho = -0.727$ vs.\ $\rho = -0.577$). For TabICL, the \emph{tabular-dominated} unimodal condition is an exception with a positive correlation ($\rho = 0.292$, $p = 4.4 \times 10^{-2}$), which is difficult to explain and worth further investigation. The two likely multimodal conditions are negative and significant, with the unimodal baseline slightly stronger ($\rho = -0.438$ vs.\ $\rho = -0.361$). TabDPT reaches significance only in the likely multimodal MM condition ($\rho = -0.366$, $p = 1.1 \times 10^{-2}$).

\section{Limitations and future work}
Our study uses a single image encoder (ViT-B/16). Extending to architecturally distinct models, such as ResNets or larger ViTs, would test whether the findings generalize across the image encoder family. A stronger image encoder could in principle reclassify some datasets we identify as likely multimodal, further illustrating that encoder choice shapes dataset characterization.

On the tabular side, SSL-based encoders and non-ICL tabular foundation models are each represented by a single model (TabM-SSL and TARTE respectively). Broader coverage of these encoder classes would test how robust the encoder-sensitivity finding in these model families.

The positive correlation between EDS and F1 rank for TabICL in the tabular-dominated unimodal setting remains unexplained and worth further investigation.

\section{Conclusion}

We present a systematic evaluation of tabular encoder choice in the tabular-image setting, spanning simple non-learned baselines through state-of-the-art ICL tabular models. Our results demonstrate that tabular encoder choice is a critical confound in multimodal benchmarking. Rankings are not stable across encoders, and conclusions drawn from a single encoder do not generalize. In modality-dominated datasets, fusing modalities can degrade performance below the unimodal baseline. In likely multimodal datasets, the observed fusion advantage shrinks as encoder quality improves, and a simple bilinear baseline with a strong encoder matches a carefully engineered method. The context-query representation shift extends beyond TabPFNv2 to TabDPT and TabICL. Vanilla extraction should be avoided across all three ICL models. NP is the recommended extraction scheme for most cases.






\begin{ack}
    Ilia is grateful to his wife KJ and his family for everything. The co-authors thank prof. Lars Schmidt-Thieme and the ISML Lab for their support and for creating a great research environment.
    We are grateful to Maxim Borisyak, Stefan Born, Thorben Werner and Ngoc Son Le for insightful discussions, and to Yury Gorishniy for kindly answering our questions about TabM.
    We thank Jörg Striewski and Kerstin Hinze-Melching for technical and administrative support.
    We also thank the anonymous NeurIPS reviewers for their thoughtful feedback.

    This work was supported by the Information Science and Machine Learning Lab (ISMLL) at the University of Hildesheim and the Volkswagen Financial Services Data Analytics Research Center (VWFS DARC). The authors gratefully acknowledge the computing time granted by the Resource Allocation Board and provided on the supercomputer Emmy/Grete at NHR-Nord@Göttingen as part of the NHR infrastructure. The calculations for this research were conducted with computing resources under the project \texttt{nhr\_ni\_starter\_25764}.
\end{ack}

\bibliography{ISMLL}

\newpage
\begin{appendices}
\section{Extra Results}
This section presents additional results omitted from the main paper due to space constraints.

\begin{figure}[h]
    \includegraphics[width=\linewidth]{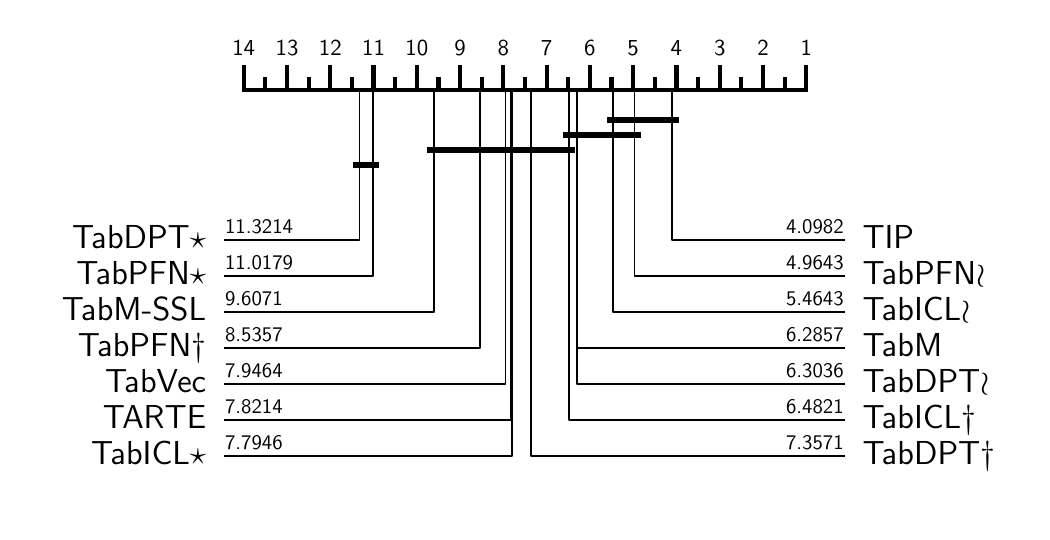}
    \caption{CDD on Multimodal Performance for all datasets}
    \label{fig:cdd_mm}
\end{figure}

\subsection{What Modality Matters?}
\label{sec:modality_contribution}
\begin{figure}[h]
  \centering
   \begin{subfigure}[b]{0.5\linewidth}
    \includegraphics[width=\linewidth]{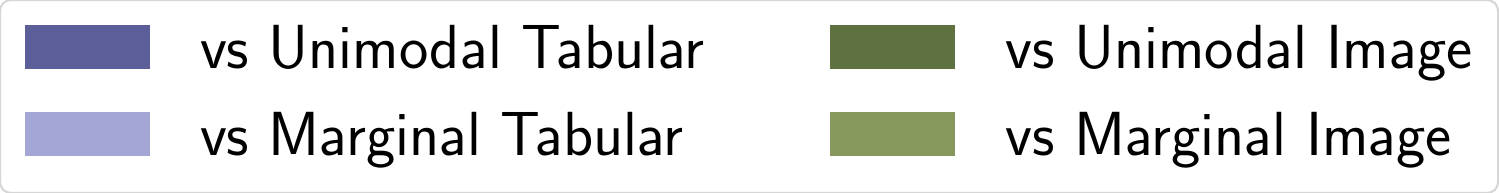}
  \end{subfigure}\\[4pt]\hfill
  \begin{subfigure}[b]{0.32\linewidth}
    \centering
    \includegraphics[width=\linewidth]{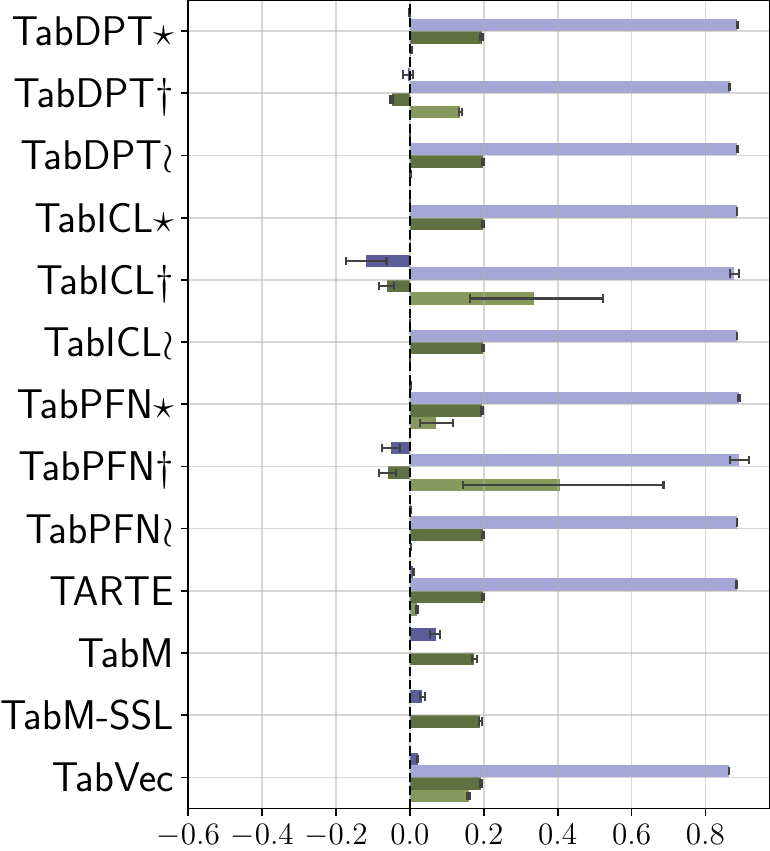}
    \caption{DVM}
  \end{subfigure}\hfill
  \begin{subfigure}[b]{0.32\linewidth}
    \centering
    \includegraphics[width=\linewidth]{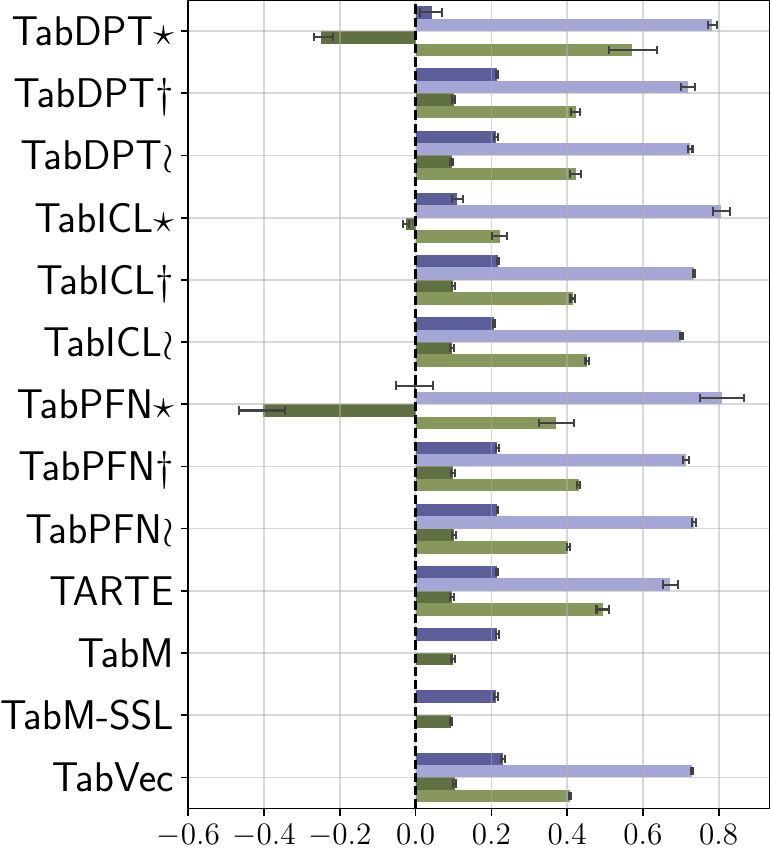}
    \caption{Wikiart}
  \end{subfigure}\hfill
  \begin{subfigure}[b]{0.32\linewidth}
    \centering
    \includegraphics[width=\linewidth]{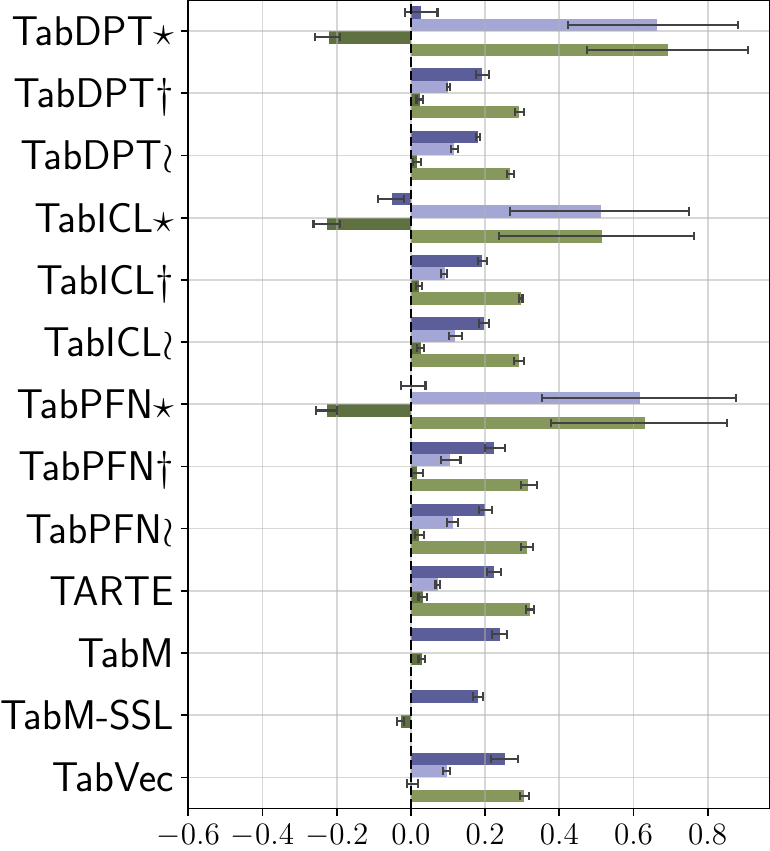}
    \caption{HAM10000}
  \end{subfigure}
  \caption{DVM (left), Wikiart (center), and HAM10000 (right).}
  \label{fig:modal_contrib}
\end{figure}

Additionally, for these three likely multimodal datasets, we quantify modality contributions more finely. Figure~\ref{fig:modal_contrib} compares multimodal performance against four baselines: the two standard unimodal models (Section~\ref{sec:exp_design}) and two Marginal-Modality estimates. For each, one modality per test instance is held fixed while the complementary modality is drawn randomly $k{=}256$ times, yielding $256n$ predictions from which F1 is computed. We refer to these as Marginal Tabular and Marginal Image depending on which modality stays intact. A value above zero indicates that the full bimodal model outperforms the marginalized single-modality estimate, confirming a genuine cross-modal interaction.

\section{Implementation Details}
This section elaborates on implementation details referenced in the main paper.

\subsection{Hyperparameter Optimization}\label{app:hps}

We report macro-averaged F1 as the primary metric. Hyperparameters are tuned with TPE~\cite{bergstraAlgorithmsHyperParameterOptimization2011} over 50 trials with 1000 sampling candidates. The hyperparameters details are shown in Table~\ref{tab:hps-both} and Table~\ref{tab:hps-tip}. Refer to the code for further details.

We use a nested cross-validation protocol with 4 inner folds and 2 outer folds. Within each outer fold, hyperparameters are selected by averaging validation F1 across the 4 inner folds. Each inner-fold model with the selected hyperparameters is then evaluated on the outer test set, yielding 4 test predictions per outer fold. No model is retrained on the full outer training set.

\begin{table}[h]
\centering
\caption{Shared hyperparameter details for Bilinear and TIP.}
\label{tab:hps-both}
\begin{tabular}{lll}
\toprule
\textbf{Hyperparameter} & \textbf{Distribution} & \textbf{Values} \\
\midrule
$d$ & Categorical & \{192, 256, 512, 768\} \\
Epochs & - & $100$ \\
Patience & - & $20$ \\
Warmup epochs & - & $10$ \\
Learning rate scheduler & - & Cos. Anneal.\ w/ Lin. Warmup \\
Optimizer & - & AdamW \\
Gradient clipping norm & - & $1$ \\
Learning rate min.\ coef. & - & $0.1$ \\
Learning rate & Log-uniform & $[10^{-5},\; 10^{-1}]$ \\
Learning rate $\gamma$ & Uniform & $[0.1,\; 1]$ \\
Weight decay & Log-uniform & $[10^{-6},\; 10^{1}]$ \\
Label smoothing & Uniform & $[0,\; 0.5]$ \\
\bottomrule
\end{tabular}
\end{table}

\begin{table}[h]
\centering
\caption{TIP-only hyperparameter details.}
\label{tab:hps-tip}
\begin{tabular}{lll}
\toprule
\textbf{Hyperparameter} & \textbf{Distribution} & \textbf{Values} \\
\midrule
Pretrain epochs & - & $250$ \\
Pretrain patience & - & $20$ \\
Pretrain learning rate & Log-uniform & $[10^{-6},\; 10^{-1}]$ \\
Frozen encoders & Categorical & \{False, True\} \\
Corruption rate & Uniform & $[0.05,\; 0.5]$ \\
CLIP temperature & Log-uniform & $[0.05,\; 1]$ \\
CLIP $\lambda_0$ & Log-uniform & $[0.05,\; 1]$ \\
Fusion dropout (attn) & Uniform & $[0,\; 0.5]$ \\
Fusion dropout (FFN) & Uniform & $[0,\; 0.5]$ \\
Encoder dropout (attn) & Uniform & $[0,\; 0.5]$ \\
Encoder dropout (FFN) & Uniform & $[0,\; 0.5]$ \\
\bottomrule
\end{tabular}
\end{table}

\end{appendices}

\end{document}